\documentclass[runningheads,a4paper]{llncs}

\usepackage{amssymb}
\usepackage{graphicx}
\usepackage{tabularx} 
\usepackage{float}
\usepackage{amsmath}
\usepackage{cleveref}

\crefname{equation}{}{}
\usepackage{color}
\usepackage{algorithm}
\usepackage{algorithmic}

\usepackage{url}
\urldef{\mailsa}\path|denis.sedov@aalto.fi, zhirong.yang@ntnu.no|
\newcommand{\keywords}[1]{\par\addvspace\baselineskip
\noindent\keywordname\enspace\ignorespaces#1}

\begin{document}

\mainmatter  


\title{Word Embedding based on Low-Rank Doubly Stochastic Matrix Decomposition}

\titlerunning{Probabilistic Word Embedding}

%
%
\author{Denis Sedov\textsuperscript{1}%
\and Zhirong Yang\textsuperscript{1,2}}

%
\authorrunning{D. Sedov and Z. Yang}

\institute{\textsuperscript{1} Department of Computer Science,
Aalto University\\
\textsuperscript{2} Department of Computer Science,
Norwegian University of Science and Technology\\
\mailsa\\
}

\def\cvs{${[}$Id: macros.tex,v 1.1 2012-02-22 07:18:00 rozyang Exp ${]}$}

\newcommand{\matA}{\mathbf{A}}
\newcommand{\matB}{\mathbf{B}}
\newcommand{\matC}{\mathbf{C}}
\newcommand{\matD}{\mathbf{D}}
\newcommand{\matE}{\mathbf{E}}
\newcommand{\matF}{\mathbf{F}}
\newcommand{\matG}{\mathbf{G}}
\newcommand{\matH}{\mathbf{H}}
\newcommand{\matI}{\mathbf{I}}
\newcommand{\matK}{\mathbf{K}}
\newcommand{\matL}{\mathbf{L}}
\newcommand{\matM}{\mathbf{M}}
\newcommand{\matN}{\mathbf{N}}
\newcommand{\matO}{\mathbf{O}}
\newcommand{\matP}{\mathbf{P}}
\newcommand{\matQ}{\mathbf{Q}}
\newcommand{\matR}{\mathbf{R}}
\newcommand{\matS}{\mathbf{S}}
\newcommand{\matT}{\mathbf{T}}
\newcommand{\matU}{\mathbf{U}}
\newcommand{\matV}{\mathbf{V}}
\newcommand{\matW}{\mathbf{W}}
\newcommand{\matX}{\mathbf{X}}
\newcommand{\matY}{\mathbf{Y}}
\newcommand{\matZ}{\mathbf{Z}}
\newcommand{\matg}{\mathbf{g}}

\newcommand{\calA}{\mathcal{A}}
\newcommand{\calB}{\mathcal{B}}
\newcommand{\calC}{\mathcal{C}}
\newcommand{\calD}{\mathcal{D}}
\newcommand{\calE}{\mathcal{E}}
\newcommand{\calF}{\mathcal{F}}
\newcommand{\calG}{\mathcal{G}}
\newcommand{\calH}{\mathcal{H}}
\newcommand{\calI}{\mathcal{I}}
\newcommand{\calJ}{\mathcal{J}}
\newcommand{\calK}{\mathcal{K}}
\newcommand{\calL}{\mathcal{L}}
\newcommand{\calM}{\mathcal{M}}
\newcommand{\calN}{\mathcal{N}}
\newcommand{\calO}{\mathcal{O}}
\newcommand{\calP}{\mathcal{P}}
\newcommand{\calQ}{\mathcal{Q}}
\newcommand{\calR}{\mathcal{R}}
\newcommand{\calS}{\mathcal{S}}
\newcommand{\calT}{\mathcal{T}}
\newcommand{\calU}{\mathcal{U}}
\newcommand{\calV}{\mathcal{V}}
\newcommand{\calW}{\mathcal{W}}
\newcommand{\calX}{\mathcal{X}}
\newcommand{\calY}{\mathcal{Y}}
\newcommand{\calZ}{\mathcal{Z}}

\newcommand{\bbA}{\mathbb{A}}
\newcommand{\bbB}{\mathbb{B}}
\newcommand{\bbR}{\mathbb{R}}
\newcommand{\bbZ}{\mathbb{Z}}
\newcommand{\bbE}{\mathbb{E}}
\newcommand{\bbH}{\mathbb{H}}

\newcommand{\veca}{\mathbf{a}}
\newcommand{\vecb}{\mathbf{b}}
\newcommand{\vecc}{\mathbf{c}}
\newcommand{\vecd}{\mathbf{d}}
\newcommand{\vece}{\mathbf{e}}
\newcommand{\vecf}{\mathbf{f}}
\newcommand{\vecg}{\mathbf{g}}
\newcommand{\vech}{\mathbf{h}}
\newcommand{\veci}{\mathbf{i}}
\newcommand{\vecj}{\mathbf{j}}
\newcommand{\veck}{\mathbf{k}}
\newcommand{\vecl}{\mathbf{l}}
\newcommand{\vecm}{\mathbf{m}}
\newcommand{\vecn}{\mathbf{n}}
\newcommand{\veco}{\mathbf{o}}
\newcommand{\vecp}{\mathbf{p}}
\newcommand{\vecq}{\mathbf{q}}
\newcommand{\vecr}{\mathbf{r}}
\newcommand{\vecs}{\mathbf{s}}
\newcommand{\vect}{\mathbf{t}}
\newcommand{\vecu}{\mathbf{u}}
\newcommand{\vecv}{\mathbf{v}}
\newcommand{\vecw}{\mathbf{w}}
\newcommand{\vecx}{\mathbf{x}}
\newcommand{\vecy}{\mathbf{y}}
\newcommand{\vecz}{\mathbf{z}}

\newcommand{\vecalpha}{\boldsymbol{\alpha}}
\newcommand{\vecbeta}{\boldsymbol{\beta}}
\newcommand{\veceta}{\boldsymbol{\eta}}
\newcommand{\vectheta}{\boldsymbol{\theta}}
\newcommand{\vecphi}{\boldsymbol{\phi}}
\newcommand{\vecpsi}{\boldsymbol{\psi}}
\newcommand{\vecrho}{\boldsymbol{\rho}}
\newcommand{\vectau}{\boldsymbol{\tau}}
\newcommand{\vecmu}{\boldsymbol{\mu}}
\newcommand{\veceps}{\boldsymbol{\epsilon}}
\newcommand{\vecxi}{\boldsymbol{\xi}}
\newcommand{\vecPhi}{\boldsymbol{\Phi}}
\newcommand{\vecDelta}{\boldsymbol{\Delta}}

\newcommand{\matDelta}{\boldsymbol{\Delta}}
\newcommand{\matEta}{\boldsymbol{\eta}}
\newcommand{\matOmega}{\boldsymbol{\Omega}}
\newcommand{\matPhi}{\boldsymbol{\Phi}}
\newcommand{\matPsi}{\boldsymbol{\Psi}}
\newcommand{\matTheta}{\boldsymbol{\Theta}}
\newcommand{\matLambda}{\boldsymbol{\Lambda}}
\newcommand{\matSigma}{\boldsymbol{\Sigma}}
\newcommand{\matzero}{\mathbf{0}}
\newcommand{\IndexSetI}{\mathcal{I}}
\newcommand{\grad}{\mathcal{\nabla}}

\newcommand{\vecone}{\mathbf{1}}
\newcommand{\veczero}{\mathbf{0}}

\def\maximize{\mathop{{\mathgroup\symoperators maximize}}}
\def\Maximize{\mathop{{\mathgroup\symoperators Maximize}}}
\def\minimize{\mathop{{\mathgroup\symoperators minimize}}}

\def\approach{\mathop{{\mathgroup\symoperators \longrightarrow}}}
\def\defineoperator{\mathop{{\mathgroup\symoperators =}}}
\newcommand{\define}{\defineoperator^{\text{def}}}

\newcommand{\Tr}{\text{Tr}}
\newcommand{\trace}{\text{trace}}
\newcommand{\diag}{\text{diag}}
\newcommand{\gradWJ}{\nabla_{\scriptscriptstyle{\matW}}\calJ}
\newcommand{\const}{\text{constant}}
\newcommand{\fracpartial}[2]{\frac{\partial #1}{\partial  #2}}

\newcommand{\defeq}{\stackrel{\text{def}}{=}}
\newcommand{\alert}[1]{\textcolor{red}{[#1]}}
\newcommand{\Shat}{\widehat{S}}
\newtheorem{law}{Law}
\newtheorem{lem}[law]{Lemma}
\newtheorem{prop}[law]{Proposition}
\newtheorem{theo}[law]{Theorem}
\newtheorem{coro}[law]{Corollary}
%
%

\maketitle

\begin{abstract}
	Word embedding, which encodes words into vectors, is an important starting point in natural language processing and commonly used in many text-based machine learning tasks. However, in most current word embedding approaches, the similarity in embedding space is not optimized in the learning. In this paper we propose a novel neighbor embedding method which directly learns an embedding simplex where the similarities between the mapped words are optimal in terms of minimal discrepancy to the input neighborhoods. Our method is built upon two-step random walks between words via topics and thus able to better reveal the topics among the words. Experiment results indicate that our method, compared with another existing word embedding approach, is more favorable for various queries.
\keywords{Nonnegative matrix factorization, Word embedding, Cluster analysis, Doubly stochastic}
\end{abstract}

\section{Introduction}
\label{sec:intro}
In recent years machine learning (ML) that involves text data has found many real-world applications \cite{manning,stamatatos,turian}. Each data item in these applications is a sequence of words and other tokens. Originally each word is represented by its ID. However, this is not suitable for machine learning, where most common ML algorithms admit vectors as their input. One-hot encoding is inefficient when the vocabulary is large. Therefore word embedding which finds a low-dimensional vectorial representation of words is a fundamental starting point.

A good word embedding method should respect the relations among the words. It is commonly to learn an embedding vector space where the neighborhoods of the words are approximately preserved. Two typical approaches include \emph{Word2Vec} \cite{word2vec} which maximizes the likelihood of each word given their neighbors (or in the reversed way) and \emph{GloVe} which minimizes a weighted squared loss between the input and output pairwise relations. Some variants of Word2Vec and GloVe have been proposed subsequently \cite{stergiou,ling_naacl,dingwall_glove}.

However, embeddings learned by the above approaches may not provide optimal similarities between the words. After the word vectors are obtained, their pairwise similarities require external measures such as cosine similarity, which can be suboptimal because the learning objective involves non-normalized word vectors. Moreover, the negative sampling trick in Word2Vec provides only an approximating surrogate. Theoretically it remains unknown whether the ad hoc choice of negative distribution guarantees that the original CBOW or Skip-Gram objectives are optimized or not.

In this paper we present a new nonnegative matrix factorization (NMF) method and apply it to learn vectorial representation of words. Our method factorizes the doubly stochastically constrained approximating matrix. In this way we directly optimize over the normalized word vectors and provide their optimal similarities in the embedding space in terms of least approximation discrepancy. Unlike Word2Vec, our method does not require extra stochastic approximation tricks or assumptions on negative distributions.
We test our method on two popularly used text data sets and compare it with the Word2Vec results. Our results indicate that the proposed method is often more favorable for various $k$-nearest-neighbor queries.

%
%

The remaining of the paper is organized as follows. In Section \ref{sec:review} we review the word embedding problem and two existing embedding methods. Next we present our new NMF method and show how to apply it to learn probabilistic representation of words in Section \ref{sec:dcd}. Our optimization algorithm is presented in Section \ref{sec:optim}. Experimental setting and results are presented in Section \ref{sec:exp}. Then in Section \ref{sec:concl} we conclude the work and discuss some future directions. 


\section{Brief Review of Previous Word Embedding Methods}
\label{sec:review}
A text corpus can be treated as a sequence of words and some other tokens such as punctuations. Originally each word is represented by their id in the vocabulary. Because many modern machine learning methods admit vectors as input, conventionally the word ids are converted into their one-hot encodings. That is, the $i$-th word in the $N$-sized vocabulary is represented by an $N$-dimensional vector with the $i$-entry is 1 and the others are zeros. Obviously, such one-hot encoding is inefficient when $N$ is large. A low-dimensional ($r$-dim with $r\ll N$) vector encoding, called word embedding, is needed for more efficient learning tasks.

Word embeddings should respect the proximity of words in the original sequence. A common requirement is that if two words often appear nearby, their mapped points in the embedding space should be close. On the other hand, if two words seldom co-occur in the same neighborhood, they should be placed distantly in the embedding.

One way to implement the above requirement is to maximize the likelihood of a language model. For example, the Word2Vec Skip-Gram method finds the vectors $\{w_t\}_{t=1}^N$ of the words which maximizes
\begin{align}
\calL(\{w_t\}_{t=1}^N) = \frac{1}{T}\sum_{t=1}^{T}\sum_{j \in \calN(t)} \log P\left(\text{word}_j|\text{word}_t\right) \; ,
\end{align}
where $\calN(t)$ is the neighborhood of location $t$ and the conditional likelihood is defined as
\begin{align}
P\left(\text{word}_j|\text{word}_t\right)=\frac{\exp\left(w_j^Tw_t\right)}{\sum_{i=1}^N\exp\left(w_i^Tw_t\right)}
\end{align}

Another approach is to approximately preserve the probability that a word appears in the neighborhood of another word. For example, GloVe implements the approximation by minimizing a weighted squared loss \cite{pennington2014glove}, assuming log-normal noise in the observed neighboring frequencies.

\section{Low-Rank Doubly Stochastic Matrix Decomposition}
\label{sec:dcd}
Although Word2Vec and GloVe are widely used, they do not provide a metric in the embedding space for retrieval. Cosine similarity as a conventional choice in natural language processing is often used to calculate, for example, $k$-nearest neighbors of a query in the embedding space. However, the cosine similarities between words are not optimized during the embedding learning. Therefore the retrieval based on such an external metric may not respect the original data distribution.

We observe that the mismatch arises mainly because the word vectors are not normalized in the learning objective, but they are normalized in the metric for retrieval. To overcome this problem we propose to use a new learning objective which explicitly involves the normalized word vectors. First, we employ the doubly stochasticity constraint to normalize the similarities in the embedding space, which enforces that each row or column of the output similarity matrix has unitary sum. This means each word in the embedding space has equal total similarity and denoises the imbalanced effect in the input space. Second, we find a low-rank nonnegative matrix which factorizes the doubly stochastic matrix, which significantly reduces the dimensionality of word vectors. Our optimization is based on multiplicative updates which are widely used in nonnegative matrix factorization (see Section \ref{sec:optim}).

Let $W$ be the word embedding matrix (rows as word vector codes). It has been shown that the above factorization problem can be reformulated as follows (see Theorem 1 in \cite{dcd}):
\begin{align}
\label{eq:div}
\minimize_{W\geq0}~& \calJ(W) = D(S||\Shat)\\
\label{eq:decomposition}
\text{subject to}~&~\Shat_{ij}=\sum_{k=1}^r\frac{W_{ik}W_{jk}}{\sum_vW_{vk}},\\
\label{eq:probconstraint}
&~\sum_{k=1}^rW_{ik}=1,~i=1\dots,N.
\end{align}
Here $D()$ is an information divergence measuring the discrepancy between between the input and output proximities $S$ and $\Shat$. We adopt the Kullback-Leibler divergence
\begin{align}
D(S||\Shat) =\sum_{i=1}^N\sum_{j=1}^N
\left(S_{ij}\log \frac{S_{ij}}{\Shat_{ij}}-S_{ij}+\Shat_{ij}\right)
\end{align}
because it accounts for Poisson noise and thus better for sparsity in $S$.

\begin{figure}[t]
	\begin{center}
		\includegraphics[width=\textwidth]{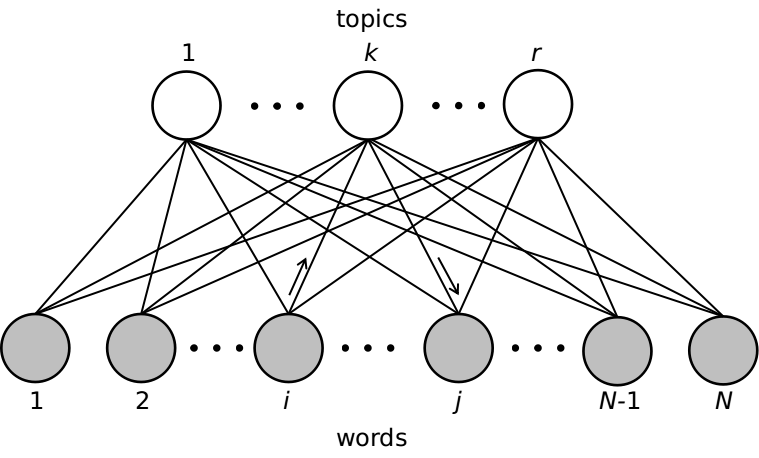}
		\caption{Word-Topic bipartite graph for $N$ words and $r$
			topics ($r<N$). The arrows show a Word-Topic-Word random walk path, which starts at the $i$th word node and ends at the $j$th word node via the $k$th
			topic node. }
		\label{fig:wtwdiag}
	\end{center}
\end{figure}

The doubly stochastic similarity matrix $\Shat$ in embedding space provides a probabilistic interpretation.
Let $W_{ik}=P\left(\text{topic}_k|\text{word}_i\right)$, the probability of assigning the $i$th data object to the $k$th topic.
Without preference to any particular word, we impose a uniform prior
$P(\text{word}_j)=1/N$ over the words. With this prior, we can compute by the Bayes' formula
\begin{align}
P(\text{word}_j|\text{topic}_k)&=\frac{P(\text{topic}_k|\text{word}_j)P(\text{word}_j)}{\sum_{v=1}^NP(\text{topic}_k|\text{word}_v)P(\text{word}_v)}\\
&=\frac{P(\text{topic}_k|\text{word}_j)}{\sum_{v=1}^NP(\text{topic}_k|\text{word}_v)}.
\end{align}
Then we can see that
\begin{align}
\Shat_{ij}=&\sum_{k=1}^r\frac{W_{ik}W_{jk}}{\sum_{v=1}^NW_{vk}}\\
=&\sum_{k=1}^r\frac{P(\text{topic}_k|\text{word}_j)}
{\sum_{v=1}^NP(\text{topic}_k|\text{word}_v)}P(\text{topic}_k|\text{word}_i)\\
=&\sum_{k=1}^rP(\text{word}_j|\text{topic}_k)P(\text{topic}_k|\text{word}_i)\\
=&P(\text{word}_j|\text{word}_i).
\end{align}
That is, if we define a bipartite graph with the words and
topics as graph nodes, $\Shat_{ij}$ is the probability that the $i$th word
node reaches the $j$th word node via a topic node (see Figure
\ref{fig:wtwdiag}). It is easy to verify that $\Shat_{ij}=\Shat_{ji}$. Therefore the output similarity matrix $\Shat$ is doubly stochastic.

\section{Optimization}
\label{sec:optim}
We implement the optimization in Eqs.~\ref{eq:div} to \ref{eq:probconstraint} by multiplicative updates.
To minimize an objective $\calJ$ over a
nonnegative matrix $W$, we first calculate the gradient and separate
it into two nonnegative parts ($\grad^+_{ik}\geq0$ and $\grad^-_{ik}\geq0$):
\begin{align}
\grad_{ik}\defeq \fracpartial{\calJ}{W_{ik}}=\grad_{ik}^+-\grad_{ik}^-.
\end{align}
Usually the separation can easily be identified from the gradient. Then
the algorithm iteratively applies a multiplicative update rule
\begin{align}
W_{ik}\leftarrow W_{ik}\frac{\grad^-_{ik}}{\grad^+_{ik}}
\end{align}
until
convergence. Such algorithms have several attractive properties, as they naturally maintain the positivity of $W$ and do not require
extra effort to tune learning step size.  For a variety of NMF
problems, such multiplicative updates monotonically decrease $\calJ$
after each iteration and therefore $W$ can converge to a stationary
point \cite{TNN2011ROZ}.

We cannot directly apply the above multiplicative fixed-point
algorithm to the proposed learning objective because there are probability constraints on the $W$
rows. Projecting the $W$ rows to the probability simplex after each
iteration would often lead
to poor local minima in practice.

Instead, we employ a relaxing strategy \cite{zhu} to handle the probability constraint. We
first introduce Lagrangian multipliers $\{\lambda_i\}_{i=1}^N$ for the constraints:
\begin{align}
\calL(W,\lambda)=\calJ(W)+\sum_i\lambda_i\left(\sum_{k=1}^rW_{ik}-1\right).
\end{align}
This suggests a preliminary multiplicative
update rule for $W$:
\begin{align}
\label{eq:mulpre}
W'_{ik} = W_{ik}\frac{\grad^-_{ik}-\lambda_i}{\grad^+_{ik}},
\end{align}
where
\begin{align}
\fracpartial{\calJ}{W}=\underbrace{\left[\left(W^TZW\right)_{kk}s_k^{-2}\right]}_{\grad^+_{ik}}
-\underbrace{\left[2\left(ZW\right)_{ik}s_k^{-1}\right]}_{\grad^-_{ik}},
\end{align}
with $Z_{ij}=S_{ij}/\Shat_{ij}$ and $s_k=\sum_{v=1}^NW_{vk}$.
Imposing $\sum_kW'_{ik}=1$ and isolating $\lambda_i$, we obtain 
\begin{align}
\lambda_i =\frac{b_i-1}{a_i},
\end{align}
where
\begin{align}
a_i=\sum_{l=1}^r\frac{W_{il}}{\grad^+_{il}},\text{ and },b_i=\sum_{l=1}^rW_{il}\frac{\grad^-_{il}}{\grad^+_{il}}.
\end{align}
Putting this $\lambda$ back in Eq.~\ref{eq:mulpre}, we obtain
\begin{align}
W_{ik}\leftarrow
W_{ik}\displaystyle\frac{\grad^-_{ik}a_i+1-b_i}{\grad^+_{ik}a_i}.
\end{align}
To maintain the positivity of $W$, we add $b_i$ to both the numerator and
denominator, which does not change the fixed point and gives the
ultimate update rule:
\begin{align}
W_{ik}\leftarrow W_{ik}\frac{\grad^-_{ik}a_i+1}{\grad^+_{ik}a_i+b_i}.
\end{align}
The above calculation steps are summarized in Algorithm
\ref{alg:wtw}. In implementation, one does not need to construct the
whole matrix $\Shat$. The ratio $Z_{ij}=S_{ij}/\Shat_{ij}$ only requires
calculation on the non-zero entries of $S$.

The above algorithm obeys a monotonicity guarantee provided by
the following theorem.
\begin{theo}
	\label{theo:monotonic}
	Denote $W^\text{new}$ the updated matrix after each iteration of Algorithm \ref{alg:wtw}.
	It holds that $\calL(W^\text{new},\lambda)\leq\calL(W,\lambda)$ with $\lambda_i = (b_i-1)/a_i$.
\end{theo}
The proof follows the Majorization-Minimization procedure
\cite{hunter2004mm,TNN2011ROZ,mmme} and is a direct corollary of Theorem 2 in \cite{dcd}.
The theorem shows that Algorithm
\ref{alg:wtw} jointly minimizes the approximation error and drives the
rows of $W$ towards the probability simplex.  The Lagrangian
multipliers are adaptively and automatically selected by the algorithm,
without extra human tuning effort. The quantities $b_i$ are the row
sums of the unconstrained multiplicative learning result, while the
quantities $a_i$ balance between the gradient learning force and the
probability simplex attraction. Besides convenience, we find that this
relaxation strategy works more robustly than the brute-force
projection after each iteration.

\begin{algorithm}[t]
	\caption{Optimization algorithm of our method}
	\label{alg:wtw}
	\begin{algorithmic}
		\STATE {\bfseries Input:} input similarity matrix $S$, number of topics $r$, positive initial guess of $W$.
		\STATE {\bfseries Output:} word embedding matrix $W$ (rows as word vectors).
		\REPEAT
		\STATE $\displaystyle \Shat_{ij}=\sum_{k=1}^r\frac{W_{ik}W_{jk}}{\sum_vW_{vk}}$
		\STATE $\displaystyle Z_{ij}= S_{ij}/\Shat_{ij}$
		\vspace{2mm}
		\STATE $s_k=\sum_{v=1}^NW_{vk}$
		\vspace{2mm}
		\STATE $\displaystyle\grad^-_{ik}= 2\left(ZW\right)_{ik}s_k^{-1}$
		\vspace{2mm}
		\STATE $\displaystyle\grad^+_{ik}= \left(W^TZW\right)_{kk}s_k^{-2}$
		\vspace{2mm}
		\STATE $\displaystyle a_i=\sum_{l=1}^r\frac{W_{il}}{\grad^+_{il}}$,~~
		$\displaystyle b_i=\sum_{l=1}^rW_{il}\frac{\grad^-_{il}}{\grad^+_{il}}$
		\vspace{2mm}
		\STATE $\displaystyle W_{ik}\leftarrow W_{ik}\frac{\grad^-_{ik}a_i+1}{\grad^+_{ik}a_i+b_i}$
		\vspace{1mm}
		\UNTIL{$W$ converges under the given tolerance}
	\end{algorithmic}
\end{algorithm}

\section{Experiments}
\label{sec:exp}

To compare the performance of our method with Word2Vec, we train word embeddings on two publicly available datasets and then construct $k$-nearest neighbor tables for specific word queries. Finally, we perform qualitative analysis on these tables and show that our method is better at capturing semantic relation between the words. The codes used in the experiments are available online\footnote{\url{https://users.aalto.fi/~sedovd1/Matrix_decomp_WE/}}.

Both training datasets used during the experiments represent a collection of English Wikipedia articles. The first dataset is \texttt{WikiText-2} \cite{merity}, which consists of 2.5M tokens with 33K words in the vocabulary. We also include \texttt{text8} dataset\footnote{\url{http://mattmahoney.net/dc/textdata.html}}, which is almost 7 times larger than \texttt{WikiText-2} and consists of 17M tokens and 254K words in the vocabulary. During the experiments, we used only top 20K most frequent words for both datasets.

For fair comparison, we followed the same default setting in the original version of Word2Vec\footnote{\url{https://github.com/tmikolov/word2vec}}. Both methods are trained on the same set of vocabulary words, the word embedding dimension is set to 200 and the size of word neighborhood equals to 8.

The results for \texttt{WikiText-2} dataset are presented in Table \ref{tab:wt2}. We can see that sometimes word neighbors produced by Word2Vec are not close semantically to the query words, whereas our method was able to produce much better results. For example, for the word ``asteroid'' the proposed method produces words like ``planets'', ``orbit'' and ``spacecraft'', which are all related to cosmos, whereas Word2Vec yields words like ``molecule'', ``insect'' and ``orientation'' that share very little in common. Moreover, it shows that the performance of Word2Vec can be quite poor for small datasets.

\begin{table}[t]
	\centering
	\caption{Seven nearest neighbors for \texttt{WikiText-2} dataset using: (top) our method and (bottom) Word2Vec.} \label{tab:wt2}
	\begin{tabularx}{\textwidth}{|l|X|}
	\hline
	word & neighbors \\
	\hline
      camera & footage, shots, shooting, screen, shoot, setting, showing \\
         zoo & gorillas, bars, spiders, exhibit, Pattycake, sharing, lowland \\
  literature & literary, poets, languages, language, writings, tradition, references \\
        moon & observations, observation, Venus, solar, transit, measurements, atmosphere \\
        coin & coins, dollar, dollars, Mint, purchase, fund, costs \\
     leather & silk, cloth, wrapped, manufactured, mud, synthetic, mills \\
        cold & warm, heat, exposed, hot, winter, falling, dry \\
      spring & winter, summer, kept, fall, brief, Over, arrival \\
       queen & ruler, mentions, kings, supreme, throne, kingdom, succession \\
    asteroid & planets, probe, orbit, spacecraft, NASA, Solar, orbits \\
	\hline
	\end{tabularx}
	\newline
	\vspace{2mm}
	\newline
	\begin{tabularx}{\textwidth}{|l|X|}
	\hline
	word & neighbors \\
	\hline
      camera & reggae, synthesizers, backup, retro, boots, carriage, bouncing \\
         zoo & griffin, Avis, Reader, headpiece, earthworks, Sleat, Owl \\
  literature & Kannada, writings, Fu, tradition, poets, Vaishnava, historical \\
        moon & skeletal, reactivity, equilibrium, sodium, spectral, infrared, triangular \\
        coin & dollar, convention, solution, price, potential, program, annular \\
     leather & twigs, pipes, longitudinal, bags, triangular, tapered, gum \\
        cold & curved, snow, reaches, beak, rough, bars, loop \\
      spring & autumn, 1900s, 1940s, 1880s, 1870s, 2000s, 1800s \\
       queen & Blanche, Wentworth, diplomat, bodyguard, mathematician, relates, prince \\
    asteroid & molecule, insect, orientation, isotope, triangle, undirected, flash \\

	\hline
	\end{tabularx}
\end{table}

Table \ref{tab:text8} shows the results for \texttt{text8} dataset. We can see that the increase in text corpus size helps to obtain more meaningful embeddings. However, Word2Vec tends to produce rather rare and specific neighbors for the query words, whereas our method produces more common words. For example in case of Word2Vec, the closest neighbors to the word ``dracula'' are ``stoker'' and ``bram'', which constitute the name of the author, who wrote the corresponding novel, as well as ``lugosi'' and ``bela'', which are related to the name of the actor portraying Dracula. On the other hand, by using our method, the close neighbors consist of the words ``frankenstein'' and ``godzilla''. These three words constitute the group of the iconic horror movie monsters and are well associated with each other.  

\begin{table}[t]
	\centering
	\caption{Seven nearest neighbors for \texttt{text8} dataset using: (top) our method and (bottom) Word2Vec.} \label{tab:text8}
	\begin{tabularx}{\textwidth}{|l|X|}
	\hline
	word & neighbors \\
	\hline
       green & blue, red, white, yellow, black, color, brown \\
     airport & railway, rail, downtown, airlines, traffic, train, metropolitan \\
   celebrity & interviews, credits, kids, favorite, talent, joy, charity \\
   microsoft & windows, operating, apple, mac, os, dos, macintosh \\
     ancient & greek, middle, historical, pre, latin, medieval, tradition \\
   byzantine & emperors, dynasty, ottoman, conquered, constantinople, conquest, rulers \\
       roman & empire, church, catholic, holy, eastern, christian, ancient \\
    dinosaur & dinosaurs, prehistoric, fossils, habitat, insect, specimen, elephants \\
     dracula & frankenstein, vampire, noir, adaptations, godzilla, cyberpunk, horror \\
    godzilla & sequel, monsters, monster, adventure, anime, horror, robot \\

	\hline
	\end{tabularx}
	\newline
	\vspace{2mm}
	\newline
	\begin{tabularx}{\textwidth}{|l|X|}
	\hline
	word & neighbors \\
	\hline
	   green & shade, lantern, purple, onion, violet, herring, panther \\
     airport & heathrow, ferry, destinations, monorail, airline, flights, hub \\
   celebrity & britney, quiz, vh, listings, portrayals, futurama, syndicated \\
   microsoft & novell, xp, excel, hypercard, borland, netscape, macromedia \\
     ancient & hellenistic, etruscan, sumerian, vedic, mycenaean, phoenician, hellenic \\
   byzantine & achaemenid, seleucid, assyrian, justinian, hittite, heraclius, frankish \\
       roman & byzantine, frankish, aztec, claudian, gaius, aurelius, seleucid \\
    dinosaur & mammal, reptiles, lizard, dodo, zebra, bipedal, skeleton \\
     dracula & stoker, bram, vampire, lugosi, bela, poirot, remake \\
    godzilla & lugosi, toho, miniseries, bela, remake, akira, highlander \\

	\hline
	\end{tabularx}
	\vspace{-3mm}
\end{table}

\section{Conclusion}
\label{sec:concl}
We have proposed a new word embedding method which is based on low-rank decomposition of doubly stochastic similarity matrix in the embedding space. Unlike previous approaches, our method provides not only the low-dimensional word vectors but also their pairwise similarity metric for subsequent applications such as retrieval. The resulting similarities are explicitly optimized in terms of least Kullback-Leibler divergence to the input similarity matrix. We have proposed an optimization algorithm based on multiplicative updates for minimizing the presented cost function.
Experiment results have shown that our method works better for two selected text corpora compared to the state-of-the-art word embedding method in terms of providing more meaningful $k$-nearest neighbors in the embedding space. 

There are several future directions. In this work we have used a batch-mode optimization algorithm, which could be replaced by using distributed and stochastic learning techniques, for example
co-distillation \cite{anil2018large}, towards a more scalable and more efficient method. We could also incorporate Bayesian treatment of the embedding vectors, for example, using Dirichlet priors \cite{sinkkonen,dcd} and automatic rank determination \cite{ardpnmf}. Moreover, the discrepancy between input and output proximity matrices could be replaced by other learnable information divergence \cite{learndiv}.
In addition, our methods is ready to be applied in other domains, for example, finding the embedding vectors of $k$-mers in DNA sequences.


\section*{Acknowledgment} 
The work is supported by Finnish Academy (grant numbers 307929 and 314177) and the Telenor-NTNU AI Lab project.

\bibliographystyle{splncs03}
\bibliography{dcd_we}

\end{document}